\documentclass{article} 
\usepackage{iclr2026/iclr2026_conference,times}

\usepackage{amsmath,amsfonts,bm}

\def\eqref#1{equation~\ref{#1}}

\def\1{\bm{1}}

\DeclareMathAlphabet{\mathsfit}{\encodingdefault}{\sfdefault}{m}{sl}
\SetMathAlphabet{\mathsfit}{bold}{\encodingdefault}{\sfdefault}{bx}{n}

\usepackage{hyperref}
\usepackage{url}
\usepackage{float}
\usepackage{graphicx}
\usepackage{subcaption}

\title{Detecting labeling bias using Influence functions }

\author{Frida Jørgensen\thanks{This work was performed at the Technical University of Denmark as part of a Bachelor’s thesis.} \\
{Copenhagen University}\\
\texttt{frida.marie@live.dk} \\
\And
Nina Weng \& Siavash Bigdeli \\
{Technical University of Denmark} \\
\texttt{\{ninwe,sarbi\}@dtu.dk} 
}

\iclrfinalcopy 
\begin{document}

\maketitle

\begin{abstract}
Labeling bias arises during data collection due to resource limitations or unconscious bias, leading to unequal label error rates across subgroups or misrepresentation of subgroup prevalence. Most fairness constraints assume training labels reflect the true distribution, rendering them ineffective when labeling bias is present; leaving a challenging question, that \textit{how can we detect such labeling bias?} In this work, we investigate whether influence functions can be used to detect labeling bias. 
Influence functions estimate how much each training sample affects a model's predictions by leveraging the gradient and Hessian of the loss function -- when labeling errors occur, influence functions can identify wrongly labeled samples in the training set, revealing the underlying failure mode.
We develop a sample valuation pipeline and test it first on the MNIST dataset, then scaled to the more complex CheXpert medical imaging dataset. 
To examine label noise, we introduced controlled errors by flipping 20\% of the labels for one class in the dataset.
Using a diagonal Hessian approximation, we demonstrated promising results, successfully detecting nearly 90\% of mislabeled samples in MNIST. On CheXpert, mislabeled samples 
consistently exhibit higher influence scores. These results highlight 
the potential of influence functions for identifying label errors.

\end{abstract}

\section{Introduction}

Most fairness constraints focus on equalizing model predictions across subgroups, i.e. fairness in prediction level, implicitly treating the training data as a faithful representation of the true distribution~\citep{blum2019recovering}. 
However, 
the assumption -- training data reflect the true underlying distribution -- breaks down \textit{when bias is introduced during data collection}. 
Labeling bias, where certain subgroups are systematically mislabeled due to resource limitations or annotator bias, embeds errors directly into the training data that prediction-level constraints cannot address~\citep{jiang2020identifying,liao2023social,favier2023fair}. \citet{blum2019recovering} showed that under labeling bias, fairness criteria including Demographic Parity, Equalized Odds, and Calibration not only fail to correct the bias but can lead to worse performance across most subgroups.

This leads to a critical question: \textit{how can we detect such labeling bias}, which is often very hidden in the data collection process?\footnote{There are methods targeting mitigating labeling bias without knowing what went wrong, which is out of the focus of this paper as we focus on detecting labeling bias first then mitigating afterwards for transparency.}

Tracing the failure mode or model debugging is usually connected with 
model explainability. If we can explain the model behavior, that might 
lead to a clue of what went wrong. 
However, much of the previous model explainability work, especially the post-doc method and the majority of in-training methods, 
focusing on explanations in prediction level, i.e. explaining predictions in relation to model parameters or test inputs. While useful, this perspective overlooks a critical factor: the role of the training data, where labeling bias is introduced. 
To truly understand why a model learns specific parameters - and thus why it makes specific predictions - one should consider the model as a function of its training data. 
This shift in perspective leads to the concept of sample valuation~\citep{xiong2024towards}, which \textit{aims to explain predictions in terms of which of the training samples are most responsible for them}. In this approach, the explanation lies not only in the model’s architecture or input features, but also in the contribution of individual training data points to the learned behavior of the model.

In this work, we investigate how sample valuation can help detect 
labeling bias. Specifically, we apply \textit{Influence Functions} 
(IF)~\citep{hampel1986robust} for sample valuation, where influence 
scores of training samples are estimated by leveraging the gradient 
and Hessian of the loss function. IF provides a computationally 
efficient alternative to other sample valuation methods such as 
Leave-One-Out retraining~\citep{Bates_2023}, which is prohibitively 
expensive due to the need for extensive model retraining. 
\citet{koh2020understandingblackboxpredictionsinfluence} introduced 
a closed-form implementation of the influence function method for 
sample valuation, which serves as the foundation for our approach.

We propose an experimental pipeline for sample valuation that enables 
evaluation of labeling bias at both the sample and group levels. At the sample level, given a misclassified test sample, the pipeline identifies the most influencial training samples to its prediction. At the group level, the pipeline computes the average influence score for each training sample across all misclassified test samples, revealing the subset of training samples that might have label errors. We validate the pipeline on two datasets: the handwritten digit dataset 
MNIST~\citep{lecun1998mnist} and the chest X-ray dataset 
CheXpert~\citep{chexpert2019}, applying label flipping to simulate 
real-world labeling bias.

We summarize the main contributions of this work as follows:
\begin{itemize}
    \item We apply Influence Function to detect labeling bias, 
    a critical fairness issue in real-world scenarios, by identifying 
    the most influential training samples to mispredicted test samples.
    
    \item We propose and validate an experimental pipeline for detecting labeling 
    bias by sample valuation on two datasets: MNIST and CheXpert.
    
    \item We demonstrate that thresholding influence scores across 
    training samples enables the detection of mislabeled samples, 
    highlighting the potential of influence functions for identifying 
    label errors.
\end{itemize}

The rest of the paper is structured as follows. 
Sec.~\ref{02_theory_related_work} covers the preliminary theory of influence function 
underlying this work. 
Sec.~\ref{sec:methodology} details the proposed sample valuation pipeline and implementation. 
Sec.~\ref{sec:exp_result} presents the experimental setup along with quantitative and qualitative results for both datasets. We discuss the results in Sec.~\ref{05_Discussion} and conclude in Sec.~\ref{sec:conclusion}.
\section{Preliminary: Influence Functions} \label{02_theory_related_work}
This section provides the theoretical preliminaries for influence 
functions, laying the groundwork for understanding the methodology 
and experiments that follow.
\subsection{Theory of Influence Function}
Influence functions are an analytical technique to determine sample influence, which is firstly discussed as a classic technique from robust statistics~\citep{hampel1974influence}. We are interested in finding out how much the   model parameters are influenced if we remove or increase a sample. In other words, the goal is to find the \textit{rate of change of model parameters} with respect to a change in the weight of a specific training sample \( j \).

\citet{koh2020understandingblackboxpredictionsinfluence}~fundamentally  demonstrated how influence functions can be applied to interpret models previously thought of as black-box models. 
Specifically, given the optimal parameters \( \theta^* \),

\begin{equation} 
    \theta^* = \arg \min_{\theta} \mathbb{E}_i[L(x_i, \theta)]
\end{equation}

where, \( \theta^* \) represents the model parameters that minimize the expected loss \( \mathbb{E}_i[L(x_i, \theta)] \) across all data points \( x_i \) in the training set. This is the standard optimization used to train the model.

To increase the influence of a specific training sample, the objective becomes:

\begin{equation}
     \hat{\theta} = \arg \min_{\theta} \mathbb{E}_i[L(x_i, \theta)] + \epsilon_j L(x_j, \theta)
\end{equation}

That represents the model parameters when the weight of a specific data point \( x_j \) is slightly increased by \( \epsilon_j \) (also called uplifting). This perturbation in the weight allows us to observe how sensitive the model parameters are to individual samples.

For an optimal model, the first-order optimality condition requires 
that the gradient of the objective with respect to the parameters 
equals zero:
\begin{equation}
\nabla \mathbb{E}_i[L(x_i, \hat{\theta})] + \epsilon_j \nabla L(x_j, \hat{\theta}) = 0
\end{equation}

To solve for the change in \( \theta \) with respect to \( \epsilon_j \), the influence function approach uses a second-order Taylor expansion around the original optimum \( \theta^* \). This approximates the perturbed solution \( \hat{\theta} \) in terms of \( \theta^* \), making use of the second derivative (Hessian) of the loss function:

\begin{equation}
\frac{d \theta}{d \epsilon_j} \approx \left( \nabla^2 \mathbb{E}_i[L(x_i, \theta^*)] \right)^{-1} \nabla L(x_j, \theta^*),
\end{equation}

where \(\left( \nabla^2 \mathbb{E}_i[L(x_i, \theta^*)] \right)^{-1}\) is equivalent to the inverse Hessian of the model, representing the second derivative of the expected loss with respect to the model parameters and \(\nabla L(x_j, \theta^*)\) is the gradient of the loss function with respect to the model parameters at sample \( j \), evaluated at \( \theta^* \). 

In order to examine sample-to-sample influence and see how influential \( x_j \) is on the model's performance on \( x_k \) we need to quantify how the loss for another sample \( x_k \) changes when the weight of \( x_j \) is increased. 
For this, chain rule is used: $
\frac{d L_k}{d \epsilon_j} = \frac{d L_k}{d \theta} \cdot \frac{d \theta}{d \epsilon_j}
$, where \( L_k = L(x_k, \theta) \) is the loss at sample \( x_k \).
Substituting the expression for \( \frac{d \theta}{d \epsilon_j} \), we get:

\begin{equation}
\frac{d L_k}{d \epsilon_j} \approx -\nabla L(x_k, \theta^*) \left( \nabla^2 \mathbb{E}_i[L(x_i, \theta^*)] \right)^{-1} \nabla L(x_j, \theta^*)
\end{equation}

This final expression, \( \frac{d L_k}{d \epsilon_j} \), represents how the loss at sample \( x_k \) is affected by an increase in the weight of sample \( x_j \). The minus sign explicitly captures the idea that increasing the weight of a sample \(\epsilon_j\) causes an update to \(\theta\) in the direction that minimizes the loss for \(x_j\), which can inversely affect the loss for \(x_k\). This influence score provides insight into which training samples have the most significant impact on the model's behavior for other samples, which is the theory that will be used in the valuation pipeline.

\subsection{Applying Influence Function in Neural Network}
The main drawback of influence functions lies in the computational challenge of calculating the exact inverse Hessian. In \citet{koh2020understandingblackboxpredictionsinfluence} the approximation method inverse-hessian-vector-product (IHVP) was used, which avoids explicitly computing the Hessian. Instead, it leverages the Hessian-vector product, making the computation as efficient as working with first-order information. However, this approach comes with a trade-off: while it scales better, some information is lost. While \citet{koh2020understandingblackboxpredictionsinfluence} produced insights for small models, and also attempted to prove their scalability other works have found that influence functions using IHVP are not necessarily accurate when scaling \citep{basu2021influencefunctionsdeeplearning}. Various alternatives to IHVP have been developed to make these calculations more scalable. FastIF~\citep{guo2021fastif}, enhances the efficiency of IFs by integrating a k-Nearest Neighbors (kNN) approach to limit the subset of candidate samples, significantly speeding up the inverse Hessian-vector product computation. A more recent paper~\citep{grosse2023studyinglargelanguagemodel} uses the Eigenvalue-Corrected Kronecker-Factored Approximate Curvature (EK-FAC) method to replace the Hessian with an approximate curvature matrix, which still somewhat preserves influence accuracy while accelerating computations. These approach demonstrates that effective influence estimation is also possible with other Hessian approximations.

\subsection{How to Interpret Influence Scores}
Positive influence scores represent training samples that increase the test sample's loss when included in training. These samples are \textit{harmful} to the decision-making process, effectively steering the model away from correctly classifying the test sample. 
Contrary, negative influence scores suggest that a training sample reduces the test sample's loss, reinforcing the model’s correct classification. These samples are considered \textit{helpful}, supporting the model’s ability to generalize.

\begin{figure}[tb]
    \centering
    \includegraphics[width=0.8\textwidth]{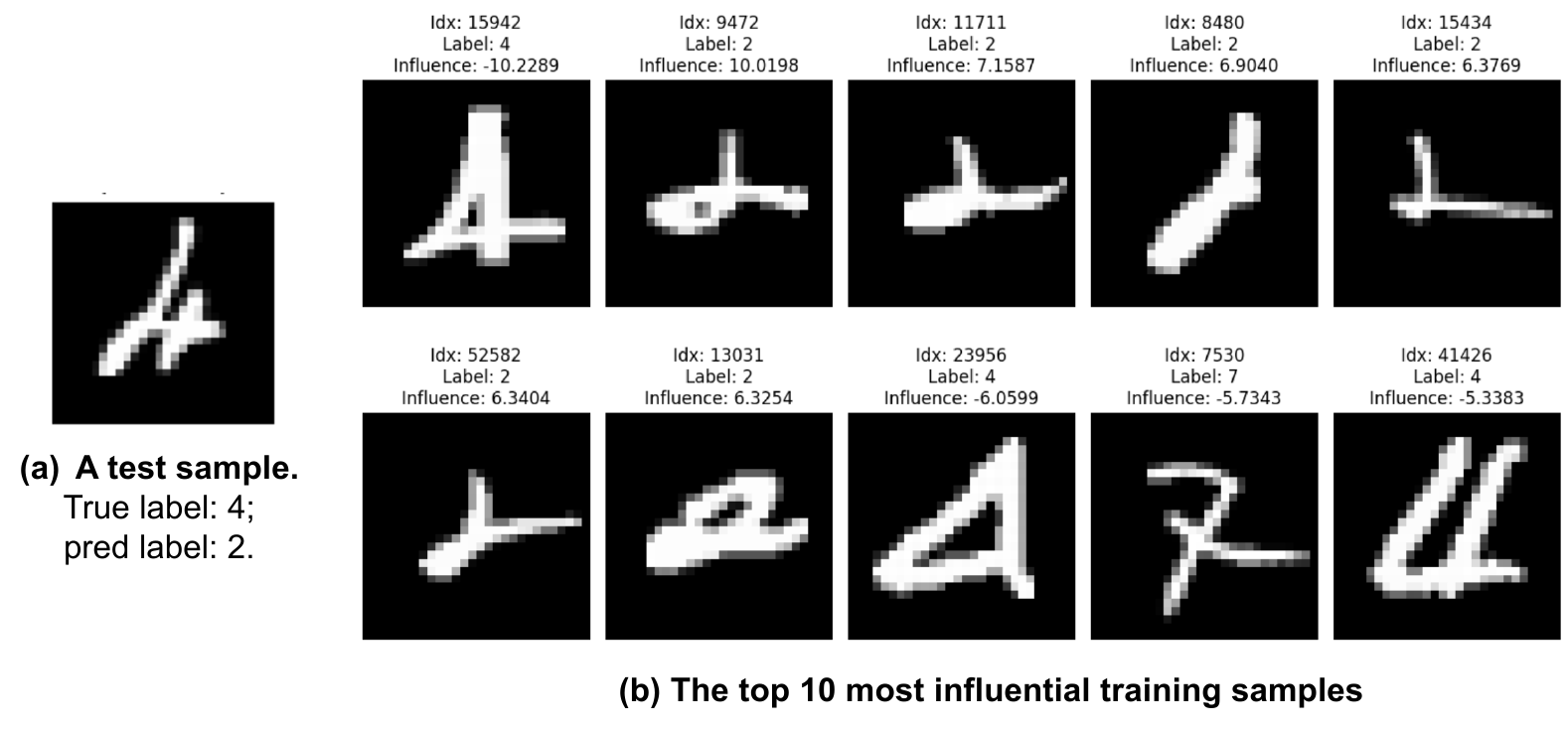}
    \caption{A misclassified test point with true label '4' and predicted label '2' along with the 10 most influential training points. The CNN Model is trained with the original MNIST.}
    \label{fig:mis_4}
\end{figure}

To illustrate,  we demonstrated an example from MNIST where a test sample is misclassified as '2' with true label of '4' (see Fig.~\ref{fig:mis_4}).
Training points with positive IF score (indexes 9472, 11711, 8480, 15434, 52582, and 13031)  are '2' samples and share similar visual traits with the misclassified test point.
These samples confused the model and are \textit{harmful} to the model’s decision process indicating by their high IF score. 
While, training points with negative IF score (indexes 15942, 23956 and 41426) share the correct label `4' and reinforce the correct classification, aligning with the expected behavior of influence scores.

\section{Methodology} \label{sec:methodology}

\begin{figure}[t]
    \centering
    \includegraphics[width=1.0\textwidth]{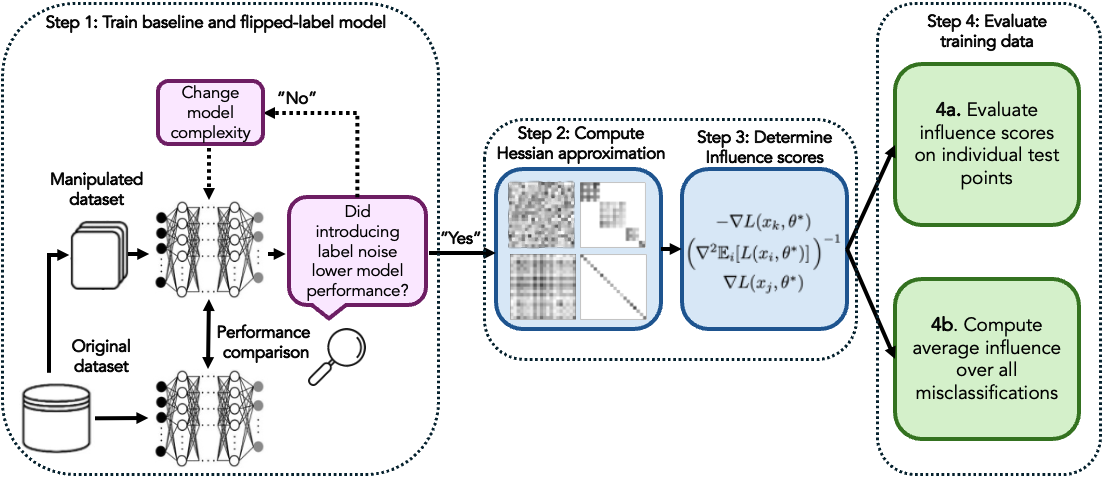}
    \caption{\textbf{Illustration of the sample valuation pipeline}: Four steps are included: (1) A baseline model and a model trained on a manipulated dataset with flipped labels are trained and sensitivity to label noise is verified (2) a Hessian-based approximation is computed (3) influence scores determined to (4) evaluate the impact of training samples on model predictions. Note that the baseline model trained on the original dataset is used solely for comparison, while all subsequent sample valuation is performed using the model trained on the manipulated dataset.}
    \label{fig:method}
\end{figure}

We proposed a sample valuation pipeline consisting of four steps (see Fig.~\ref{fig:method}). 
First, we train both a baseline model and a model on a manipulated dataset with flipped labels, verifying that the model is sensitive to label noise. Second, we compute a Hessian-based approximation to enable efficient influence score calculation. Third, we calculate influence scores for each training sample with respect to misclassified test samples. Finally, we evaluate the impact of training samples on model predictions, identifying potentially mislabeled samples based on their influence scores.
The following sections detail model selection, Hessian approximation, and influence score calculation.

\subsection{Model selection}
The primary objective of model selection across all experiments was to ensure that the chosen architectures were meaningfully affected by labeling bias.
Since this work investigates whether influence functions can detect harmful mislabeled data, it is essential that the experimental setup first establishes a clear and measurable negative impact of label noise on model learning and generalization.

To this end, multiple architectures were evaluated during preliminary experiments. 
For MNIST, a three-layer convolutional model showed the clearest sensitivity to flipped labels and was therefore selected. 
When trained on noisy labels, this model exhibited substantially higher validation loss and lower accuracy compared to the baseline, confirming its sensitivity to label noise.

For CheXpert, the higher complexity of medical X-ray images required a deeper architecture. ResNet18 was chosen as it is well-established for image classification tasks and provides sufficient representational capacity to model complex medical data. The choice also enabled evaluation of the scalability of the influence-function-based sample valuation pipeline in deep neural networks.

\subsection{Hessian Approximation}
An important consideration for many applications of the Hessian is the efficiency, with which it can be evaluated. If there are $W$ parameters  
, then the Hessian matrix has dimensions $W \times W$, resulting the computational effort of evaluating the Hessian scale to $O(W^2)$ for each pattern in the data set. This becomes unfeasible for deep neural networks, so an approximation method is needed.

Using only the diagonal of the full Hessian offers a practical and efficient alternative. Just like HPV,  this approach dramatically reduces computational cost and avoids storing the full matrix. The number of computational steps required to evaluate this approximation drops to $O(W)$ with diagonal Hessian, compared with $O(W^2)$ for the full Hessian, making it feasible even for large-scale models, while avoiding some of the fragility associated with the implicit HPV approximation method.  

To compute the diagonal of the Hessian,  we use the \texttt{Laplace}~\citep{laplace2021} toolbox.
For MNIST,  
Hessian structure is set to \texttt{diag}, configuring the Laplace approximation to estimate only the diagonal elements.
For CheXpert, \texttt{subset\_of\_weights} is set to \texttt{last\_layer} to restrict the Hessian computation to the final layer. In both cases, the resulting Hessian approximation is inverted to obtain the inverse Hessian required for computing influence scores.

\subsection{Influence Score Calculation} 

To quantify the influence of each training sample on a specific test point, a closed-form influence function is used as described in Sec.~\ref{02_theory_related_work}. 
This approach enables efficient influence score computation by combining test and training gradients with an inverse Hessian approximation.

In inference time, we calculate the gradient of the test point’s loss with respect to the model parameters. The gradient vector for the test point, then serves as the reference for assessing the influence of each training sample. We then iterate through each batch in the training set, calculating the gradient of the loss for each individual training sample. The gradient is then scaled by the inverse Hessian diagonal approximation. And finally calculate the influence scores as the negative dot product between the test gradient vector and the vector scaled by the inverse Hessian approximation. 

\section{Experimental Set-up and Results} \label{sec:exp_result}
\subsection{Dataset and Implementation Details}
Two datasets were used: a 
handwriting digits dataset MNIST \citep{lecun1998mnist} and a medical imaging (chest x-ray) dataset CheXpert \citep{chexpert2019}. 
MNIST dataset consists of 70,000 grayscale images of handwritten digits with a resolution of $28 \times 28$, divided into 60,000 training examples and 10,000 test examples. Each image represents a single digit from 0 to 9.
CheXpert is a large-scale chest X-ray dataset and benchmark for automated chest X-ray interpretation. It contains labels for 
14 clinical observations along with radiologist-labeled reference 
standards for evaluation.

\subsubsection{Experimental set-up on MNIST}
\paragraph{Creating dataset with label noise.} The effect of label noise on model influence is investigated by introducing controlled label flips in a subset of the training data. We first trim MNIST to MNIST49, by keeping only the data labeled "4" and "9", leaving respectively 5842 and 5949 samples for training. 
20\% of samples labeled as "9" in the MNIST dataset were randomly selected and relabeled as "4," creating a noisier training dataset to examine the model's response to this label perturbation. Note that now the class "9" is representing the subgroup where more label error occurs, and the pipeline aim at detecting such wrongly-labeled samples.
 
\paragraph{Model training and influence score evaluation.} A shallow neural network model was chosen with an architecture of only three convolutional layers. This choice was driven by the nature of the task - a relatively simple binary classification problem with a small training dataset. The model was trained on both the original and modified datasets. To quantify the influence of each training point on misclassified test samples, influence scores were computed using the valuation pipeline. The influence values for training samples in the flipped label subset were then compared against those for non-flipped samples, focusing on influence scores in relation to misclassified test points. 

\subsubsection{Experimental set-up on CheXpert}

\paragraph{Creating dataset with label noise.} For  CheXpert, a binary classification problem was constructed for simplicity. One disease, "Pleural Effusion," was chosen as the target label, and the model was tasked with predicting whether the disease was present (1) or not (0). This label was selected due to its relatively balanced distribution of positive and negative cases, making it more suitable for controlled label noise experiments and facilitating clearer insights into model behavior and valuation. 
To introduce label noise, 20\% of the positive labels for "Pleural Effusion" were flipped to negative. 

\paragraph{Model training and influence score evaluation.} Two ResNet18 models, initialized with pretrained weights and trained with early stopping, were used.
One as a baseline on the training data with the original label distribution and another trained on the dataset with 20\% flipped labels. After the approximation of Hessain, influence scores were computed for each training point. The training samples were then ranked by their  influence scores to identify the most impactful training samples.

\paragraph{Further approximation of Hessian. }Given the computational complexity of calculating the Hessian for the full ResNet18 model, which contains 11 million trainable parameters, a practical adaptation was implemented. Inspired by prior work within the field of uncertainty quantification~\citep{daxberger2022laplacereduxeffortless}, the Hessian computation was restricted to the last layer of the network. In this adaptation, the last layer is treated as a stand-in for the entire model, with its input serving as the model’s effective input. Because the last layer has very few parameters, calculating the full Hessian is feasible and since it provides a more reliable approximation, the full Hessian is computed.

\subsection{Quantitative Results: detection ability of influence score}

\begin{figure}[tb]
    \centering

    \begin{subfigure}[b]{0.48\textwidth}
        \centering
        \includegraphics[width=\textwidth]{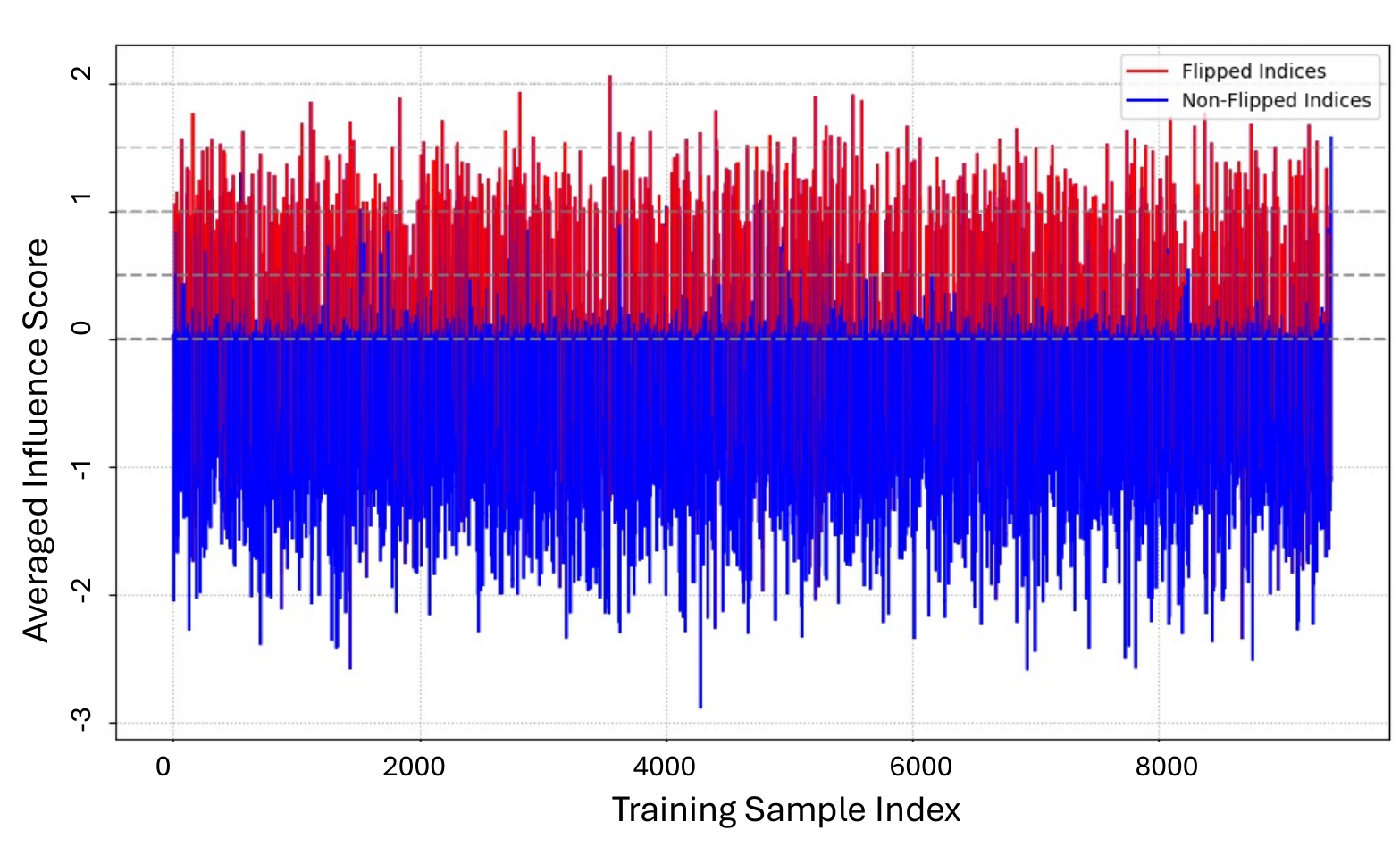}
        \caption{Average influence score per training sample on all misclassified samples (MNIST).}
        \label{fig:FL_thresh1}
    \end{subfigure}
    \hfill
    \begin{subfigure}[b]{0.48\textwidth}
        \centering
        \includegraphics[width=\textwidth]{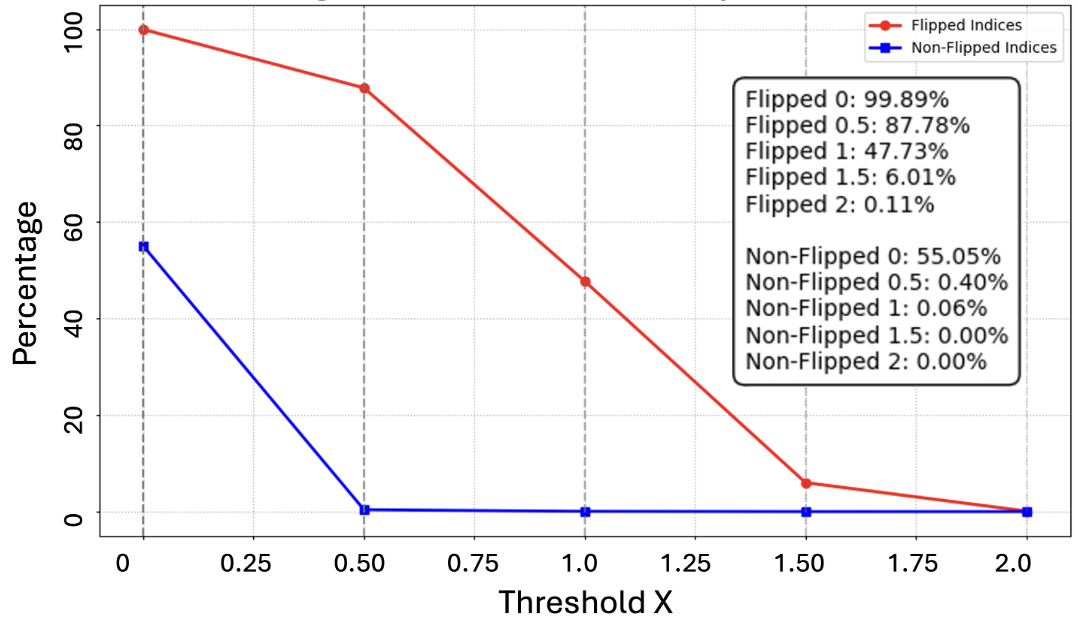
}
        \caption{Percentage of flipped samples detected at varying 
        influence thresholds (MNIST).}
        \label{fig:FL_thresh2}
    \end{subfigure}

    \vspace{1em}

    \begin{subfigure}[b]{0.48\textwidth}
        \centering
        \includegraphics[width=\textwidth]{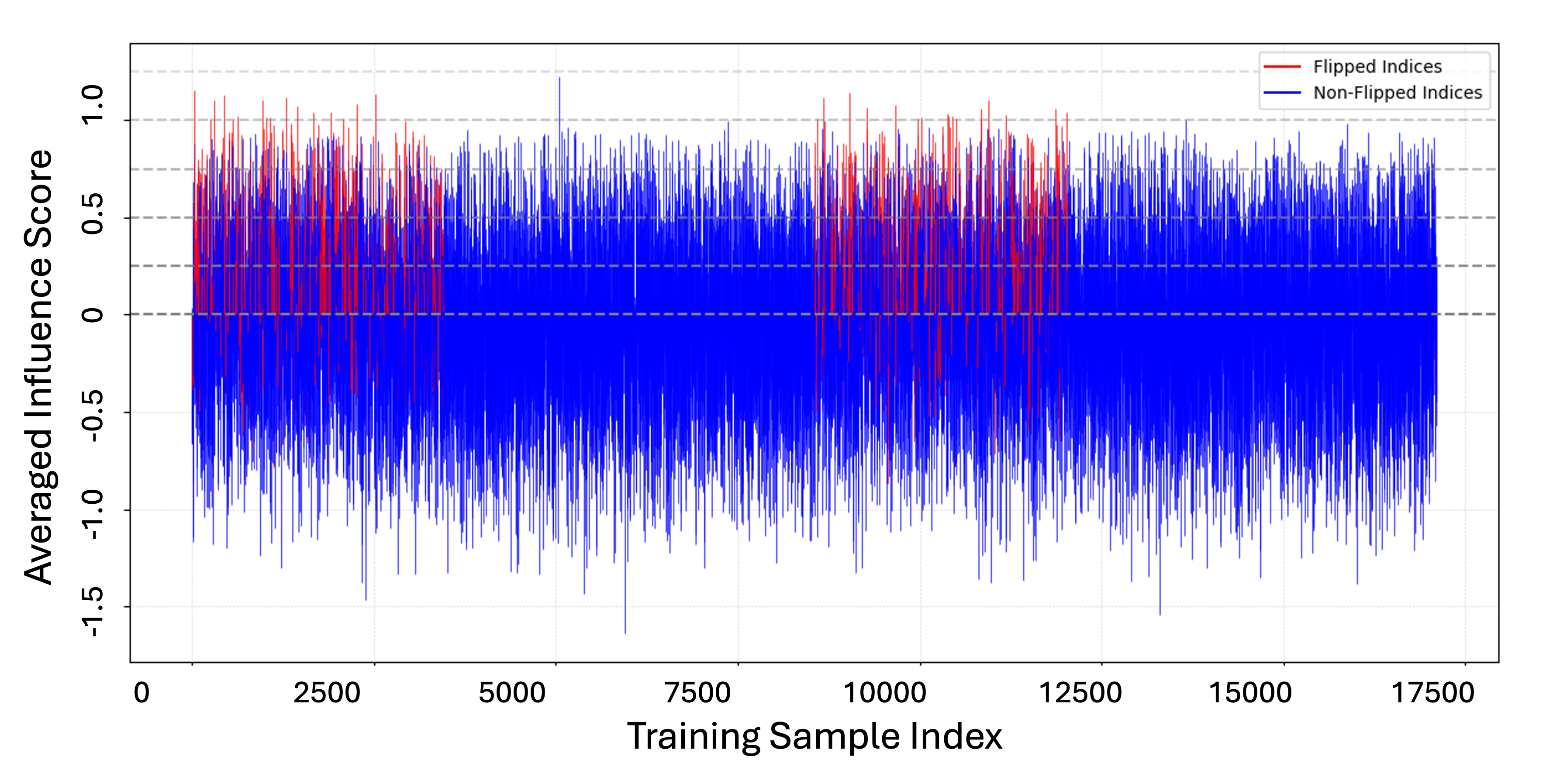}
        \caption{Average influence score per training sample on all misclassified samples (CheXpert).}
        \label{fig:FLMD_thresh1}
    \end{subfigure}
    \hfill
    \begin{subfigure}[b]{0.48\textwidth}
        \centering
        \includegraphics[width=\textwidth]{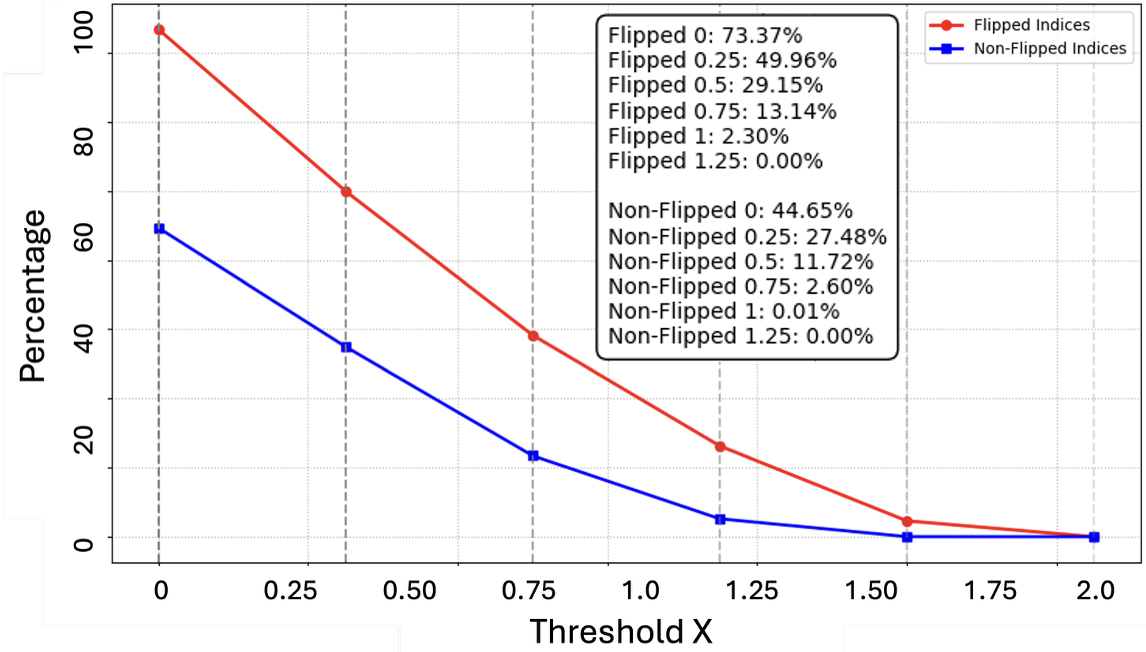}
        \caption{Percentage of flipped samples detected at varying 
        influence thresholds (CheXpert).}
        \label{fig:FLMD_thresh2}
    \end{subfigure}

    \caption{\textbf{Detection of mislabeled samples via influence scores.} 
    Left: average influence score for each training sample, where 
    higher scores indicate greater contribution to test 
    misclassifications. Right: percentage of artificially flipped 
    labels recovered at different influence score thresholds.}
    \label{fig:combined_figures}
\end{figure}

Fig.~\ref{fig:combined_figures} summarizes the influence score behavior for both the MNIST and CheXpert experiments. 
The left column (Fig.~\ref{fig:FL_thresh1} and Fig.~\ref{fig:FLMD_thresh1}) shows the average influence of each training point on misclassified test samples, such that each bar represents a training point, with its height indicating the cumulative influence score across \textit{all} misclassified test samples. The training points corresponding to flipped labels, i.e. the wrong labels, are highlighted in red, while those with correct labels are shown in blue. While the right column (Fig.~\ref{fig:FL_thresh2} and Fig.~\ref{fig:FLMD_thresh2}) reports the percentage of flipped and non-flipped indices exceeding specified positive influence thresholds.

\paragraph{MNIST}
As shown in Fig.~\ref{fig:FL_thresh1}, the majority of flipped samples exhibit high positive influence scores, indicating they are harmful to model performance. In contrast, most non-flipped samples show negative scores, meaning they reduce the loss on misclassified test points. 
Fig.~\ref{fig:FL_thresh2} shows how quickly the non-flipped (blue) detection rate drops as the threshold increases in the MNIST experiment. With just a small step up to a threshold of 0.5, the percentage of detected non-flipped points plummets to 0.40\%, while almost 90\% of the flipped labels are still detected. 

\paragraph{CheXpert}
As shown in Fig.~\ref{fig:FLMD_thresh1}, flipped samples again tend to exhibit higher positive influence scores. However, the separation between flipped and non-flipped samples is less distinct than in MNIST, as non-flipped samples also show higher positive scores. The visible clustering of flipped indices results from the original dataset structure, where Pleural Effusion positive cases were grouped rather than uniformly distributed; this did not affect training as the data loader shuffled samples. Fig.~\ref{fig:FLMD_thresh2} confirms that while flipped samples remain more prevalent at higher thresholds, the separation is weaker than in MNIST. Potential explanations are discussed in Sec.~\ref{05_Discussion}.

\subsection{Qualitative Results: the harmful and helpful samples for a misclassified example}

\begin{figure}[tb]
    \centering
    \scalebox{0.9}{%

    \begin{minipage}[c]{0.15\textwidth}
        \centering
        \includegraphics[width=\textwidth]{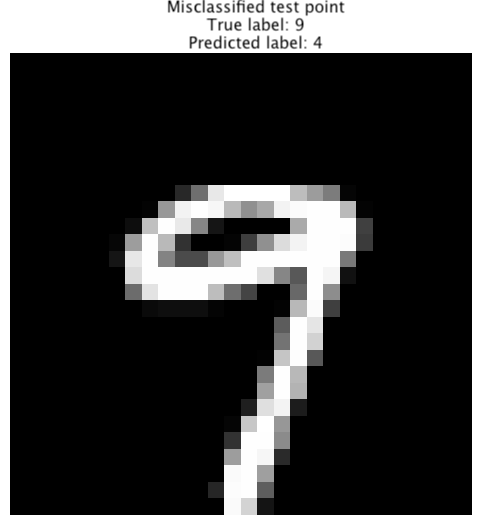}
        \subcaption{Misclassified test sample}
        \label{fig:mnist_mis_example}
    \end{minipage}
    \hfill

    \begin{minipage}[c]{0.75\textwidth}
        \centering
        \includegraphics[width=0.95\textwidth]{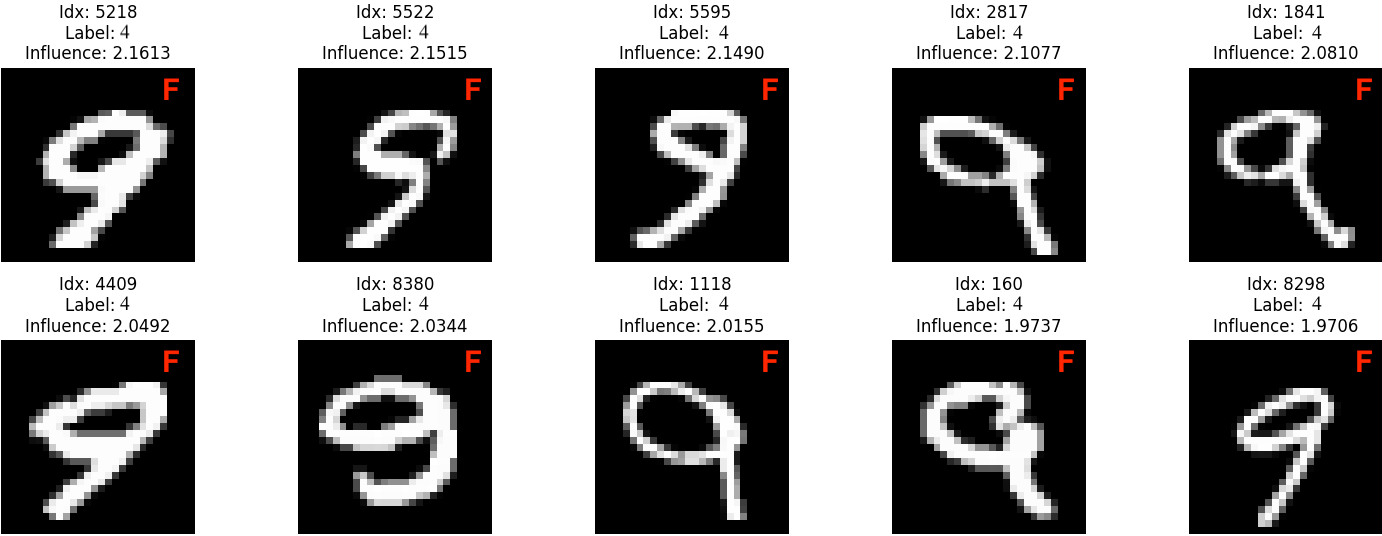}
        \subcaption{Top 10 most harmful training samples (highest positive influence)}
        \label{fig:mnist_harmful}
        
        \vspace{0.5em}
        
        \includegraphics[width=0.95\textwidth]{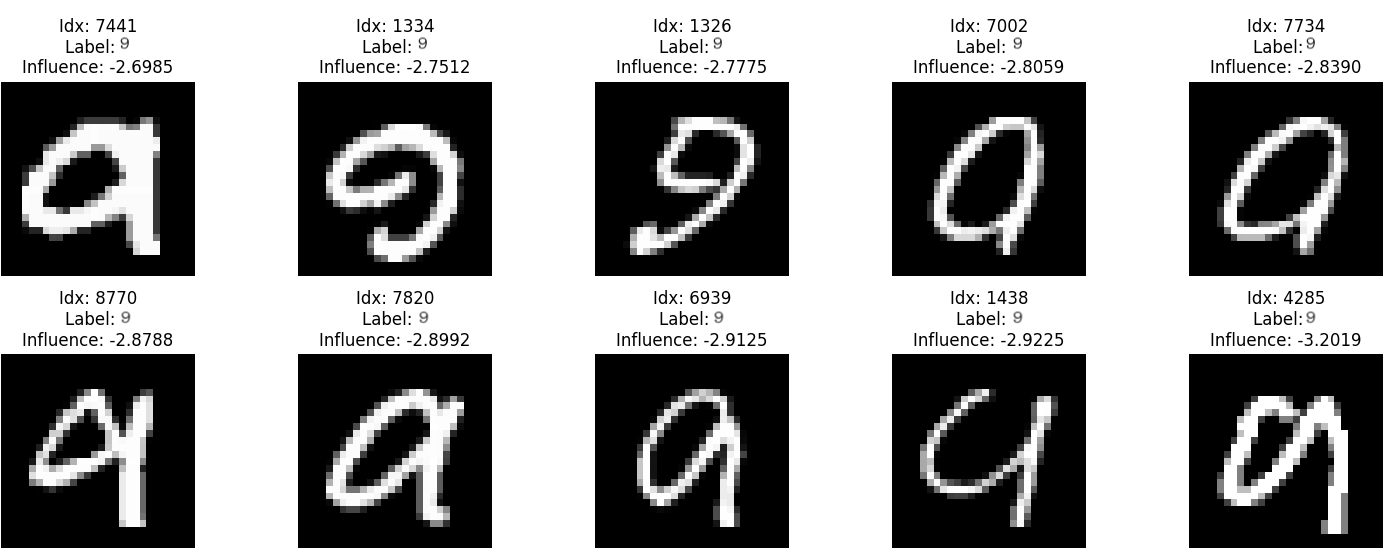}
        \subcaption{Top 10 most helpful training samples (highest negative influence)}
        \label{fig:mnist_helpful}
    \end{minipage}
    }
    
    \caption{\textbf{Visualization of a misclassified test sample and its most 
    influential training samples. }Left: a test sample misclassified by 
    the model trained on flipped labels. Right: the 10 training 
    samples with the highest positive influence (harmful) and the 10 
    with the highest negative influence (helpful) toward the correct 
    prediction. Labels shown are those used during training; samples 
    marked with `F' indicate flipped labels.}
    \label{fig:combined_example_sizes_FL}
\end{figure}

\begin{figure}[tb]
    \centering
    \scalebox{0.9}{%

    \begin{minipage}[c]{0.15\textwidth}
        \centering
        \includegraphics[width=\textwidth]{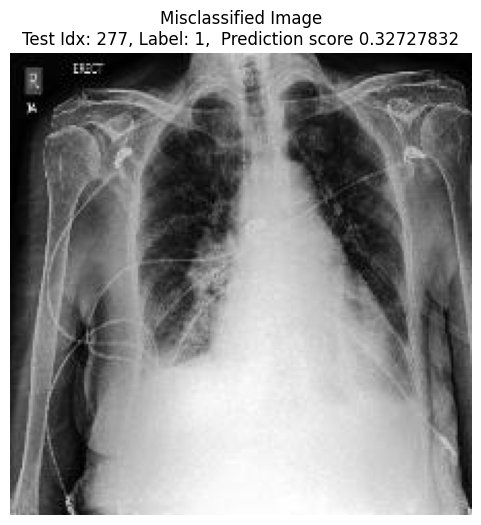}
        \subcaption{Misclassified test sample}
        \label{fig:chexpert_mis_example}
    \end{minipage}
    \hfill
 
    \begin{minipage}[c]{0.75\textwidth}
        \centering
        \includegraphics[width=\textwidth]{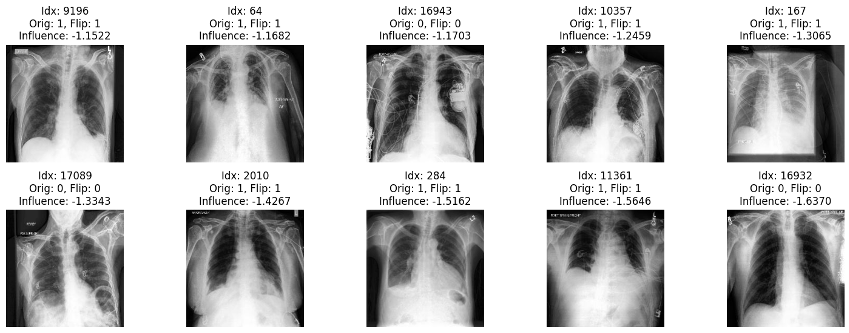}
        \subcaption{Top 10 most harmful training samples (highest positive influence)}
        \label{fig:chexpert_harmful}
        
        \vspace{0.5em}
        
        \includegraphics[width=\textwidth]{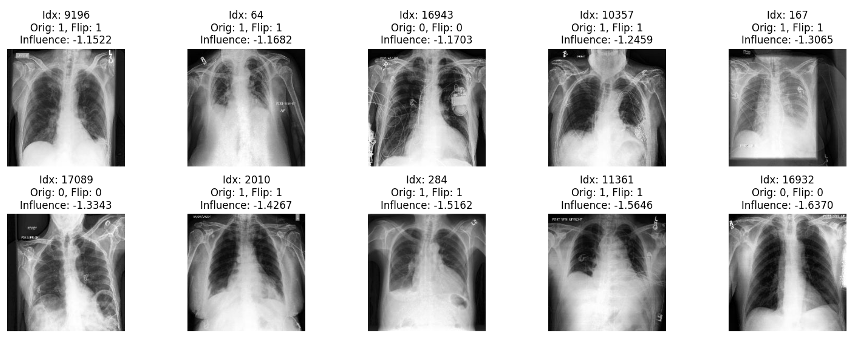}
        \subcaption{Top 10 most helpful training samples (highest negative influence)}
        \label{fig:chexpert_helpful}
    \end{minipage}
    }
    \caption{Visualization of a misclassified CheXpert test sample and 
    its most influential training samples. Left: a test sample 
    misclassified by the model trained on flipped labels. Right: the 
    10 training samples with the highest positive influence (harmful) 
    and the 10 with the highest negative influence (helpful) toward 
    the correct prediction.}
    \label{fig:combined_example}
\end{figure}

We also visualize the top 10 most harmful and helpful training samples for a given misclassified test sample. Examples for MNIST and CheXpert are shown in Fig.~\ref{fig:combined_example_sizes_FL} 
and~\ref{fig:combined_example}, respectively.

\paragraph{MNIST}
As shown in Fig.\ref{fig:combined_example_sizes_FL}, all 10 of the most harmful influential training points originate from the flipped-label distribution.
This suggests that these mislabeled training examples push the model’s predictions in a direction that increases the loss at the test point, ultimately leading to its misclassification. The 10 most helpful training points, which reduce the loss for the test point, are all '9's with correct labels. This indicates that influence scores are effective at detecting label errors.

\paragraph{CheXpert}
As shown in Fig.~\ref{fig:combined_example}, 
only three out of the ten most harmful training points come from the flipped label distribution. 
However, considering that only 20\% of training samples are flipped, the fact that flipped samples appear among the most harmful more frequently than random chance would suggest is still significant. In addition, the harmful training samples include some scans with unusual characteristics, such as odd patterns or artifacts.

\section{Discussion}
\label{05_Discussion}
\paragraph{From MNIST to CheXpert: scalability challenges.}
The MNIST experiments yielded very compelling results for using influence functions in sample valuation, and for detecting label errors. 
Can this approach be effectively extended to larger, deeper models like ResNet, where even hessian approximation becomes less feasible and potentially more imprecise?
The CheXpert experiment provided a partial answer: flipped labels again had a higher-than-random presence among the most harmfully influential training points for misclassified test samples; 
however, the separation between flipped and non-flipped labels was considerably weaker compared to which in MNIST . 

\paragraph{Possible explanations of degradation for CheXpert.}
Several factors may explain this degradation. 
First, the models might not fully learn the complex structures in chest radiographs; as uncertain and missing labels were excluded, and the dataset was further restricted to one scan per patient to avoid inaccuracies caused by labeling errors associated with multiple scans per patient.
Second, the Hessian was computed by only the last layer of the model. 
This strong approximation could reduce the reliability of the computed scores, especially when capturing the complex relationships in deeper layers of the model.

\paragraph{Outlook.}
Despite these limitations, thresholding influence scores successfully 
identified flipped labels among high-influence samples, suggesting 
potential for automated mislabel detection. Addition to that, extending the Hessian 
computation to include additional layers beyond the final one may 
improve reliability and is a promising direction for future work. 
Whether this approach generalizes to more complex, real-world 
datasets remains an open question.

\paragraph{Comparison with prior work on influence functions and fairness.}

Influence functions, as one of the sample valuation explanations, have been applied for fair machine learning in the literature. Most of which focusing on a new training strategy with IF score, given that IF score together with sensitive attributes annotation offers an chance to give some of the samples higher weighting in training for fairer performance among subgroup, e.g. re-training on a subset of weighty samples will lower the fairness violation~\citep{wang2022understanding}, reweighting samples based on IF of subgroups~\citep{li2022achieving} or sensitive attributes~\citep{wang2024fairif}, data augmentation based on IF score~\citep{xie2025accuracy}. IF scores are also used in fairness assessment, e.g. quantifying the influence of different features in a dataset on the bias of a classifier ~\citep{ghosh2023biased}. In contrast, our work applies influence functions to detect labeling bias -- a direction that, to our knowledge, remains largely unexplored.

\section{Conclusion} \label{sec:conclusion}

This work investigated the use of influence functions for detecting labeling bias in training data. On MNIST, the approach achieved strong results, detecting over 90\% of mislabeled samples with less than 1\% false positives, demonstrating clear separation between flipped and correctly labeled samples based on influence scores.

Scaling to the more complex CheXpert dataset proved challenging. While mislabeled samples still exhibited higher influence scores on average, the separation was less distinct, likely due to limited training data, model complexity, and the last-layer Hessian approximation. Despite these limitations, the results suggest that influence functions capture meaningful signal even in challenging settings.

Our findings highlight both the promise and current limitations of influence-based mislabel detection. Future work could explore extending the Hessian computation to additional layers, investigating alternative approximation strategies, and validating the approach on larger-scale medical imaging datasets. More broadly, this work contributes to the growing effort to detect and mitigate labeling bias -- a critical step toward fairer and more reliable machine learning systems.
\newpage

\bibliography{sections/reference}
\bibliographystyle{iclr2026/iclr2026_conference}

\end{document}